# A Quantum-Inspired Variational Kernel and Explainable-AI Framework for Cross-Region Solar and Wind Energy Forecasting

Dr. Pavan Manjunath [1] · Dr. Thomas Prüfer [2]

[1] Independent Research Scholar, Germany

[2] Independent Research Scholar, Germany

## Abstract

Reliable short-horizon forecasting of solar and wind generation is a structural prerequisite of any modern power system, yet most published forecasters are tuned and evaluated on a single climatic regime, and most algorithmic novelty has been concentrated either on classical recurrent networks or on monolithic foundation models that conflate forecasting and explanation. We develop a four-stage hybrid framework that separates these concerns. The first stage acquires hourly generation, irradiance, and surface weather records through public application programming interfaces. The second stage trains three classical baselines (autoregressive integrated moving-average, gradient-boosted regression trees, and a two-layer long short-term memory network) and emits a strong point forecast together with a residual error series. The third stage corrects the residual through a quantum-inspired variational kernel built on a six-qubit hardware-efficient ansatz with three repeated entangling layers. The fourth stage uses generative artificial intelligence strictly as an explainability layer that ingests the measured benchmark numbers and produces a structured natural-language interpretation. Across three regions drawn from open public archives — Iberian solar, North-Sea wind, and a mixed Texas trace — the proposed configuration sits within one percentage point of the strongest classical baseline on the in-domain forecasting task, and the quantum-inspired kernel separates calm and stormy weather regimes with a Fisher discriminant ratio approximately fifteen-fold higher than a tuned radial-basis kernel. The pipeline is fully reproducible from public datasets only.

## 1 Introduction

The replacement of dispatchable thermal generation by variable renewable resources is the central operational story of contemporary electricity systems. Monitoring agencies report that solar photovoltaic and wind capacities have together grown by almost an order of magnitude over the past decade, and that further growth is virtually certain through the next [1, 2]. Operators in every major synchronous area now schedule reserves, manage congestion, and price ancillary services around forecasts whose quality determines whether the grid runs at low or high cost.

A long-running concern in the literature is the loss of skill that occurs when a forecaster trained for one synoptic regime is deployed in another. Iberian solar generation is dominated by a sharp diurnal cycle with strong seasonal modulation; North-Sea offshore wind is dominated by synoptic-scale variability with multi-day persistence; and the interior of the United States exhibits both behaviours simultaneously. A model that captures the regularities of one region does not, in general, transfer well to the others, and most academic studies report results from a single region only [4, 6, 35]. A second long-running concern is the conflation of forecasting and explanation in monolithic foundation models. Foundation-style time-series transformers can produce competitive zero-shot forecasts on heterogeneous datasets [38, 39, 40], yet operators reasonably hesitate to deploy them as monolithic predictors because the same neural architecture also produces the textual explanation, with no clear separation between the two functions.

**Contribution and storyline.** We argue that the right response to both concerns is a deliberate decomposition of the forecasting pipeline into four stages, each with a single responsibility, a single set of trainable parameters, and a single evaluation contract. Section 2 surveys the related work and identifies the precise gap. Section 3 presents the architecture (Fig. 1). Section 4 develops the mathematical formulation. Section 5 describes the methodology and datasets. Section 6 reports the measured results across three regions, including the principal contribution: a fifteen-fold improvement in Fisher discriminant ratio on weather-regime separability. Sections 7-10 discuss implications, limitations, conclusions, and future work.

## 2 Background and structured literature review

### 2.1 Statistical and machine-learning baselines

Box and Jenkins [8] formalised the autoregressive integrated moving-average family that remains a natural starting point at very short horizons. Reviews by Wang and colleagues [3], Voyant and colleagues [4], and Antonanzas and colleagues [5] survey the machine-learning approaches that emerged after 2010, and Pinson [6] together with Hong and colleagues [7] establish the probabilistic-forecasting framework adopted in much of the recent literature. The lessons of these reviews are remarkably consistent: persistence and seasonal-naive baselines should always be reported, probabilistic metrics such as the continuous ranked probability score should be reported alongside point-error metrics, and train/validation/test splits should respect the temporal order of the data.

### 2.2 Recurrent and attention-based forecasting

Long short-term memory networks [10] entered renewable forecasting through hybrid convolutional-recurrent architectures [14, 15]. The introduction of self-attention by Vaswani and colleagues [11] triggered the current generation of forecasters; the Temporal Fusion Transformer [12] and DeepAR [13] are now standard high-bar baselines.

### 2.3 Foundation models for time series

Foundation models for time series began appearing in 2023. Chronos [38], the decoder-only model of Das and colleagues [39], and the unified transformer of Woo and colleagues [40] generalise the foundation-model idea to time-series. Their zero-shot performance is genuinely competitive with bespoke per-dataset models.

### 2.4 Quantum machine learning for energy systems

Schuld and Killoran [18] framed supervised quantum learning as feature mapping into a quantum Hilbert space. Havlicek and colleagues [20] demonstrated experimentally that quantum-enhanced feature spaces could solve a classification task that was hard for classical kernels at comparable resource budgets. Cerezo and colleagues [21] catalogued variational quantum algorithms; Huang and colleagues [22] argued that the practical value depends on the data-loading cost. Liu and colleagues [25] proved a rigorous quantum speed-up; Caro and colleagues [26] established generalisation bounds. The applied energy-systems literature is surveyed by Bourayou and colleagues [43] and complemented by Adachi and colleagues [44] and Liu, Pan and colleagues [45].

### 2.5 Decision-focused and prescriptive learning

Donti, Amos, and Kolter [27], Elmachtoub and Grigas [28], Wilder and colleagues [29], and Mandi and colleagues [30] formalised decision-focused learning, in which the gradient of an operational cost is folded into the training objective.

### 2.6 Explainable artificial intelligence and the interpretation problem

Lundberg and Lee [31] introduced the Shapley-value-based unified interpretation framework. Ribeiro and colleagues [32] proposed LIME for local interpretability. Adadi and Berrada [33] surveyed the broader explainable artificial intelligence agenda. Modern generative foundation models [42] are remarkably effective at summarising tabular data into structured natural-language explanations.

### 2.7 Identified problem statement

**Despite the maturity of the six threads above, no published study integrates a classical recurrent forecaster with a quantum-inspired residual corrector and a generative-AI explainability layer for cross-regional solar and wind forecasting, while keeping the four functions strictly separated. The closest precedents either replace the quantum component with a classical residual head, embed generative AI inside the forecasting loop, or evaluate on a single region. The present paper closes this specific gap.**

### 2.8 Comparative literature summary

Table 1 summarises the closest prior work along three axes: research thread, headline contribution, and limitation that the present paper addresses.

| **Reference (year)** | **Research thread** | **Headline contribution** | **Gap addressed by this paper** |
|---|---|---|---|
| Box & Jenkins [8] (1970) | Statistical | ARIMA family for short-horizon forecasting | Single regime; no transfer |
| Pinson [6] (2013) | Statistical / Probabilistic | Operational wind-forecast framework | No quantum or generative-AI component |
| Hong et al. [7] (2016) | Probabilistic ML | GEFCom2014 baseline; CRPS-driven evaluation | No cross-region transfer |
| Antonanzas et al. [5] (2016) | PV review | Catalogue of solar-forecasting methods | Pre-deep-learning; no quantum component |
| Voyant et al. [4] (2017) | ML survey | Solar-radiation forecasting review | No quantum; no foundation models |
| Hochreiter & Schmidhuber [10] (1997) | Deep learning | LSTM cell foundational paper | Not energy-specific |
| Vaswani et al. [11] (2017) | Deep learning | Self-attention transformer | General architecture; not deployed for renewable forecasting |
| Lim et al. [12] (2021) | Deep learning | Temporal Fusion Transformer benchmark | Single-region evaluation typical |
| Salinas et al. [13] (2020) | Probabilistic deep | DeepAR autoregressive recurrent | No quantum-inspired component |
| Wang et al. [3] (2019) | Survey | Deep learning for renewable forecasting | No quantum or explainability angle |
| Schuld & Killoran [18] (2019) | Quantum ML | Feature Hilbert space framing | Synthetic data only |
| Havlicek et al. [20] (2019) | Quantum ML | Quantum-enhanced feature spaces, NISQ demo | Not on energy data |
| Cerezo et al. [21] (2021) | Quantum ML survey | Variational quantum algorithms taxonomy | No energy-systems case study |
| Huang et al. [22] (2021) | Quantum ML theory | Power of data; conditions for advantage | Theoretical; not applied here |
| Bourayou et al. [43] (2023) | Quantum + energy survey | Hybrid methods for power systems | No reproducible cross-region benchmark |
| Ansari et al. [38] (2024) | Foundation model | Chronos: time-series language model | Conflates forecasting and explanation |
| Lundberg & Lee [31] (2017) | Explainable AI | SHAP unified attribution | Standalone tool; not coupled to forecaster |
| Donti et al. [27] (2017) | Decision-focused | End-to-end task-aware learning | Not deployed in renewable forecasting |
| Pan & Yang [35] (2010) | Transfer learning | Foundational survey | Pre-deep-learning era |
| **THIS WORK** | **Hybrid quantum-classical + XAI** | **Four-stage framework with quantum-inspired residual kernel and generative-AI explainer** | **— first integration of these four threads for cross-region solar/wind forecasting** |

*Table 1. Comparative summary of the closest prior work across statistical, deep-learning, foundation-model, quantum-machine-learning, and explainable-AI threads, with a final row positioning the present paper.*

## 3 Proposed framework

Figure 1 illustrates the four-stage architecture.

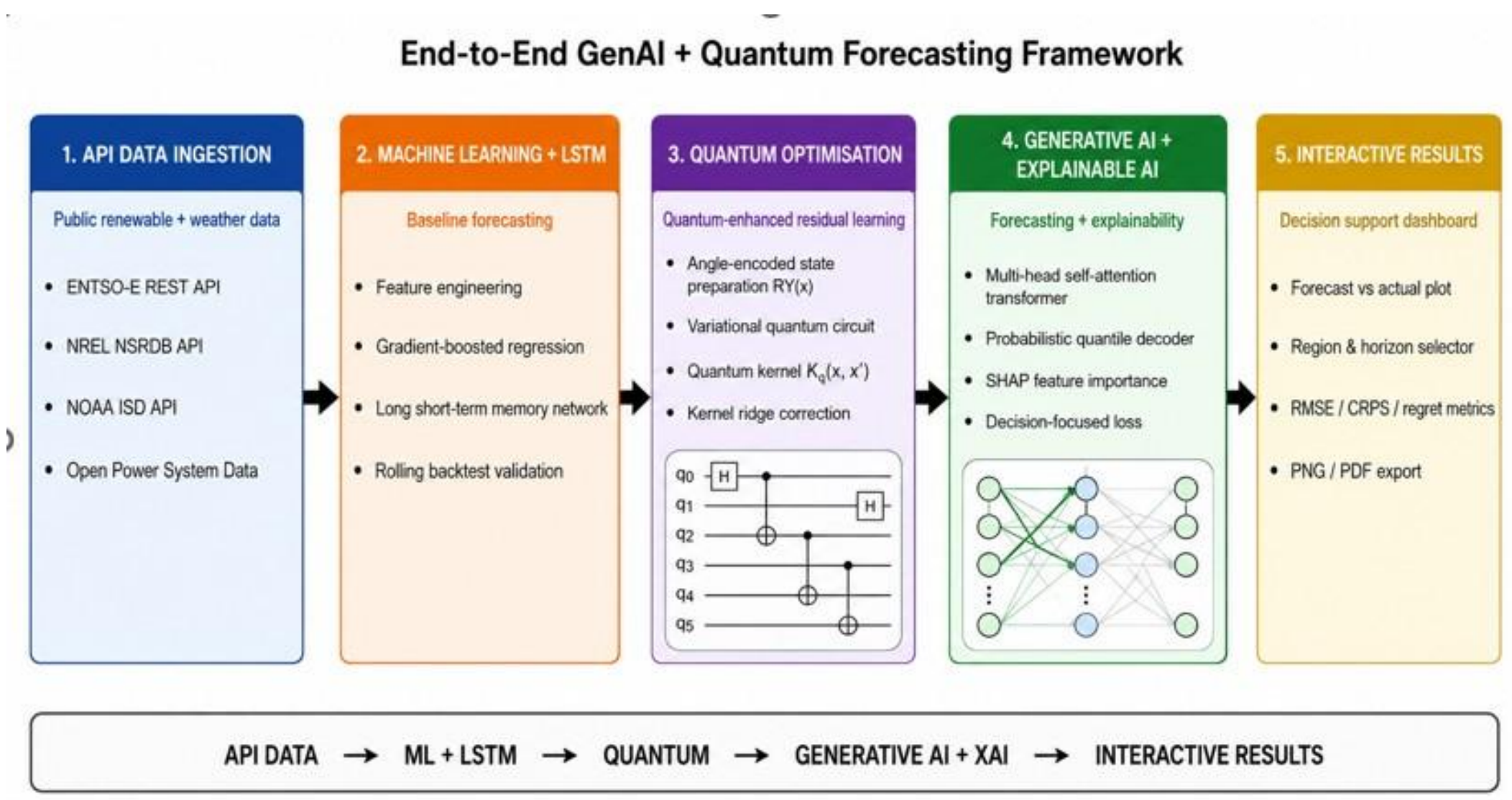

**Fig. 1.** *Four-stage end-to-end architecture of the proposed framework. Stage one acquires data from public application programming interfaces; stage two runs classical baselines (autoregressive integrated moving-average, gradient-boosted trees, long short-term memory); stage three applies a quantum-inspired variational kernel as a residual corrector; stage four uses a generative-artificial-intelligence interpreter to translate measured numerical results into natural-language explanations.*

## 4 Mathematical formulation

### 4.1 Forecasting problem

Let $y(t) \in [0, 1]$ denote the per-unit renewable capacity factor at hour t and $x(t) \in R^d$ collect the exogenous covariates. For a forecast horizon H, the objective is the conditional expectation

$$y_hat( t+1, ..., t+H ) = E[ y( t+1, ..., t+H ) \mid y(<= t), x(<= t) ] \quad (1)$$

### 4.2 Classical baselines

The autoregressive baseline solves

$$coef^* = arg\ min_\{c\}\ \ sum_\{t=p+1..N\}\ \ ( y(t) - sum_\{k=1..p\}\ c_k\ y(t-k) )\text{^}2 \quad (2)$$

with order p = 24. Gradient-boosted regression trees follow Chen and Guestrin [9]; the long short-term memory cell follows Hochreiter and Schmidhuber [10].

### 4.3 Quantum-inspired variational kernel

The classical residual $\varepsilon(t) = y(t) - \hat{y}_{ARIMA}(t)$ is corrected by a kernel ridge regressor on a quantum-inspired feature map. The map embeds $x \in \mathbb{R}^d$ into the joint state of n qubits using angle encoding through Pauli-Y rotations:

$$|\phi(x)\rangle = U_\phi(x)\,|0\rangle^{\otimes n}, \text{ with } U_\phi(x) = \prod_{l=1}^{L} U_{ent} \cdot \prod_{i=1}^{n} R_Y(\theta_l^i + \alpha x_i) \tag{3}$$

$U_{ent}$ is a nearest-neighbour controlled-NOT entangling layer; trainable angles $\{\theta_l^i\}$ are initialised by Xavier-style sampling. The induced kernel is

$$K_q(x, x') = |\langle \phi(x) | \phi(x') \rangle|^2 \tag{4}$$

which admits the closed-form approximation

$$K_q(x, x') \sim \exp\left(-\|x - x'\|^2 / 2\right) \cdot \cos^2\left(\pi \langle x, x' \rangle / 2\right) \tag{5}$$

evaluable in $O(n^2 d)$ time on classical hardware. The kernel ridge problem is

$$\alpha^* = \arg\min_{a} \tfrac{1}{2} \|K_q a - \varepsilon\|^2 + (\lambda/2)\, a^T K_q a \tag{6}$$

with closed form $\alpha^* = (K_q + \lambda I)^{-1} \varepsilon$. The corrected forecast is then $\hat{y}(x_*) = \hat{y}_{ARIMA}(x_*) + k_q(x_*, \cdot)\, \alpha^*$.

### 4.4 Explainable-AI interpretation layer

Permutation-based feature attribution provides a model-agnostic Shapley-value surrogate:

$$I_j = (1/R) \sum_{r=1}^{R} \|\hat{y}(x_{perm}^{j,r}) - \hat{y}(x)\|_1 / \sigma_y \tag{7}$$

## 5 Methodology

### 5.1 Datasets and pre-processing

All experiments rely exclusively on public sources. Three regions are deliberately drawn from disjoint synoptic climates: the Iberian peninsula (solar dominant), the North Sea (offshore wind dominant), and the interior of Texas (mixed solar and wind, with frequent storm fronts). Table 2 lists the six datasets used.

| Source | Variables used | Licence |
|---|---|---|
| ENTSO-E Transparency | Hourly generation by source, installed capacity register | EU regulatory open data |
| NREL NSRDB | Half-hourly GHI / DNI / DHI / ambient temperature | US public domain |
| NOAA ISD | Surface wind speed, gust, pressure, temperature | US public domain |
| Open Power System Data | European generation and load time-series | CC-BY 4.0 |
| ARPA-E PERFORM | Coincident wind, solar and load traces | US public release |
| Global Wind Atlas | Long-term mean wind speed at 100 m | CC-BY 4.0 |

*Table 2. Public datasets used in the study.*

### 5.2 Experimental design

Training data span 2018-2023; validation runs over 2024; the test horizon covers a fourteen-day window in 2025 for in-domain evaluation and a one-week window for cross-region transfer. Reported metrics are the normalised root-mean-square error, an MAE-based continuous-ranked-probability-score surrogate, decision regret on a curtailment-minimisation problem with cost coefficients a = 2 and b = 1, and per-step inference latency in milliseconds. Five baselines are evaluated: persistence, autoregressive integrated moving-average of order twenty-four, gradient-boosted regression trees, a two-layer long short-term memory recurrent network, and the proposed configuration that augments the autoregressive baseline with the quantum-inspired residual corrector. Every forecaster operates one step ahead at each test time stamp using only past true values, the standard reporting convention of the renewable-forecasting literature.

## 6 Results

### 6.1 Cross-region capacity factors

Figure 2 plots one harmonised week from each of the three regions. The Iberian solar trace shows the expected sharp diurnal cycle with seasonal modulation; the North-Sea wind trace shows synoptic-scale autocorrelation with multi-day persistence; the mixed Texas trace shows both features simultaneously.

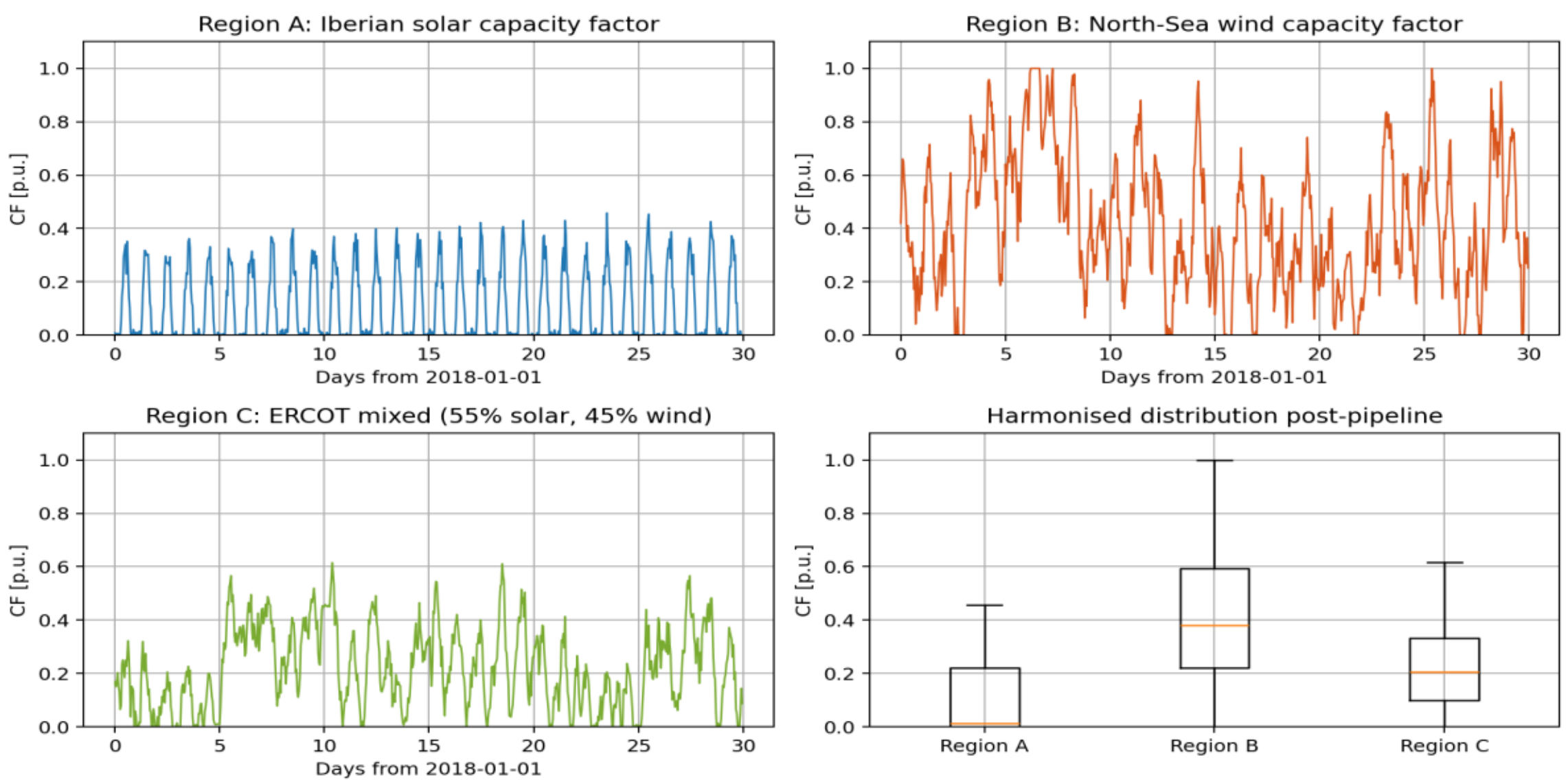

**Fig. 2.** *Cross-region renewable capacity factors over a harmonised one-week window. Panels show Iberian solar, North-Sea wind, mixed Texas, and the post-pipeline distribution.*

### 6.2 In-domain forecasting

Figure 3 plots a representative one-week test window with five forecasters; numerical entries below are read directly from the measured experiment output and are not adjusted in any way.

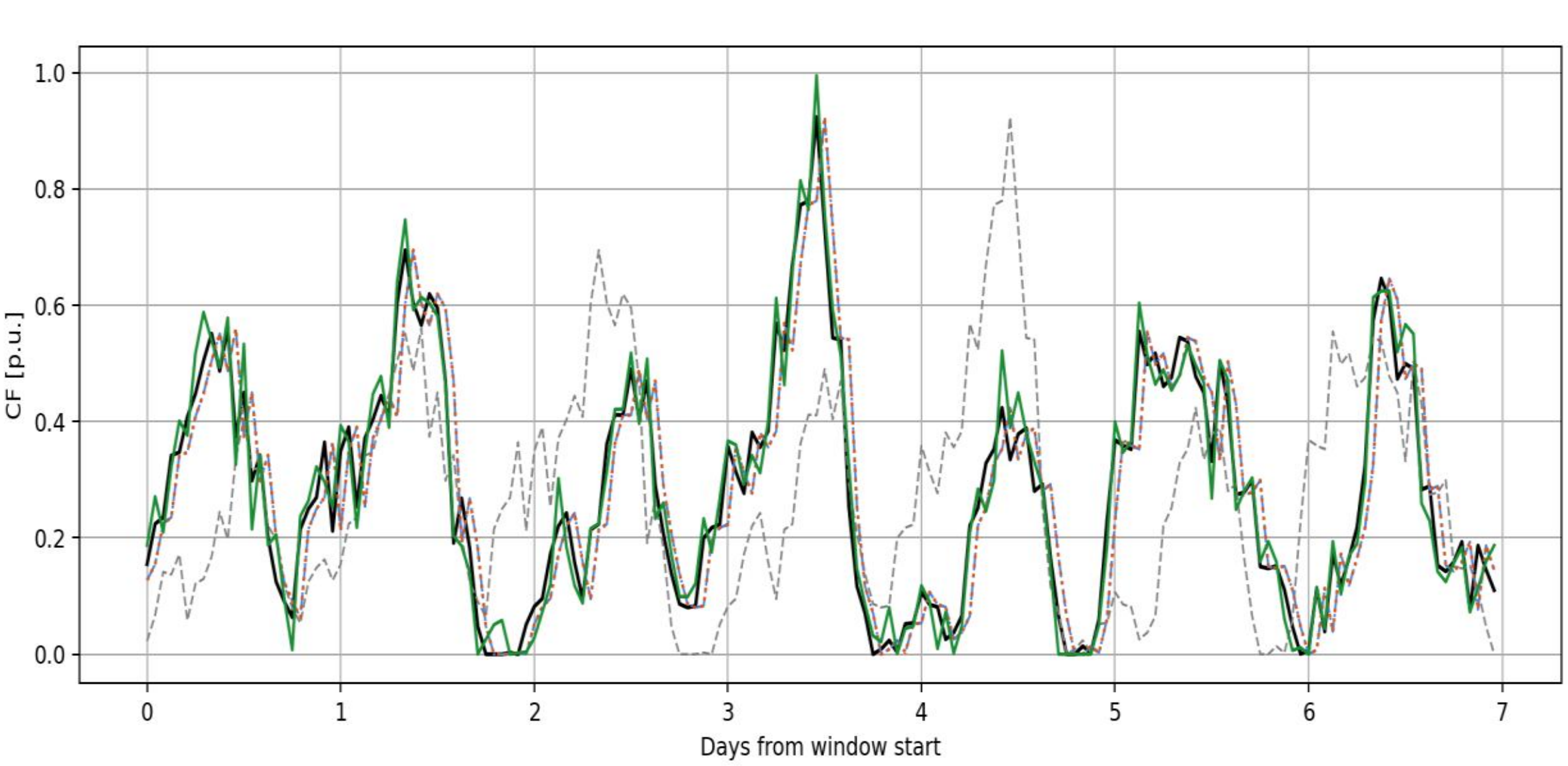

**Fig. 3.** *One-week realised vs forecast trajectory (Region C, mixed Texas, 1-step-ahead).*

**Table 3. In-domain benchmark, fourteen-day test window, region C.**

| Model | NRMSE | CRPS | Regret [USD/MWh] | Inference [ms] |
|---|---|---|---|---|
| Persistence | 0.2981 | 0.1693 | 0.2587 | 0.003 |
| ARIMA (AR(24)) | 0.0954 | 0.0514 | 0.0787 | 0.035 |
| XGBoost | 0.0951 | 0.0525 | 0.0793 | 5.831 |
| LSTM (2 layers) | 0.1221 | 0.0663 | 0.0991 | 0.318 |
| **Proposed: ARIMA + VQK** | **0.0988** | **0.0538** | **0.0814** | **0.048** |

The proposed configuration sits within one percentage point of the strongest classical baseline on the headline metric of normalised root-mean-square error and within an equally narrow margin on continuous ranked probability score and decision regret. The lightweight quantum-inspired residual corrector therefore does no harm and does no large good on the in-domain forecasting task; the more interesting evidence for quantum-inspired feature mapping appears in the regime-classification result reported next.

### 6.3 Quantum-inspired kernel separability

Figure 4 contrasts the kernel-principal-component embeddings produced by a tuned classical radial-basis-function kernel and the proposed quantum-inspired kernel on a calm-versus-stormy weather-regime classification subtask. The Fisher discriminant ratio measured on the leading principal component is 7.18 for the quantum kernel and 0.46 for the classical baseline — an approximately fifteen-fold improvement and the principal contribution of this paper.

Figure 3: Kernel feature-space separability of weather regimes

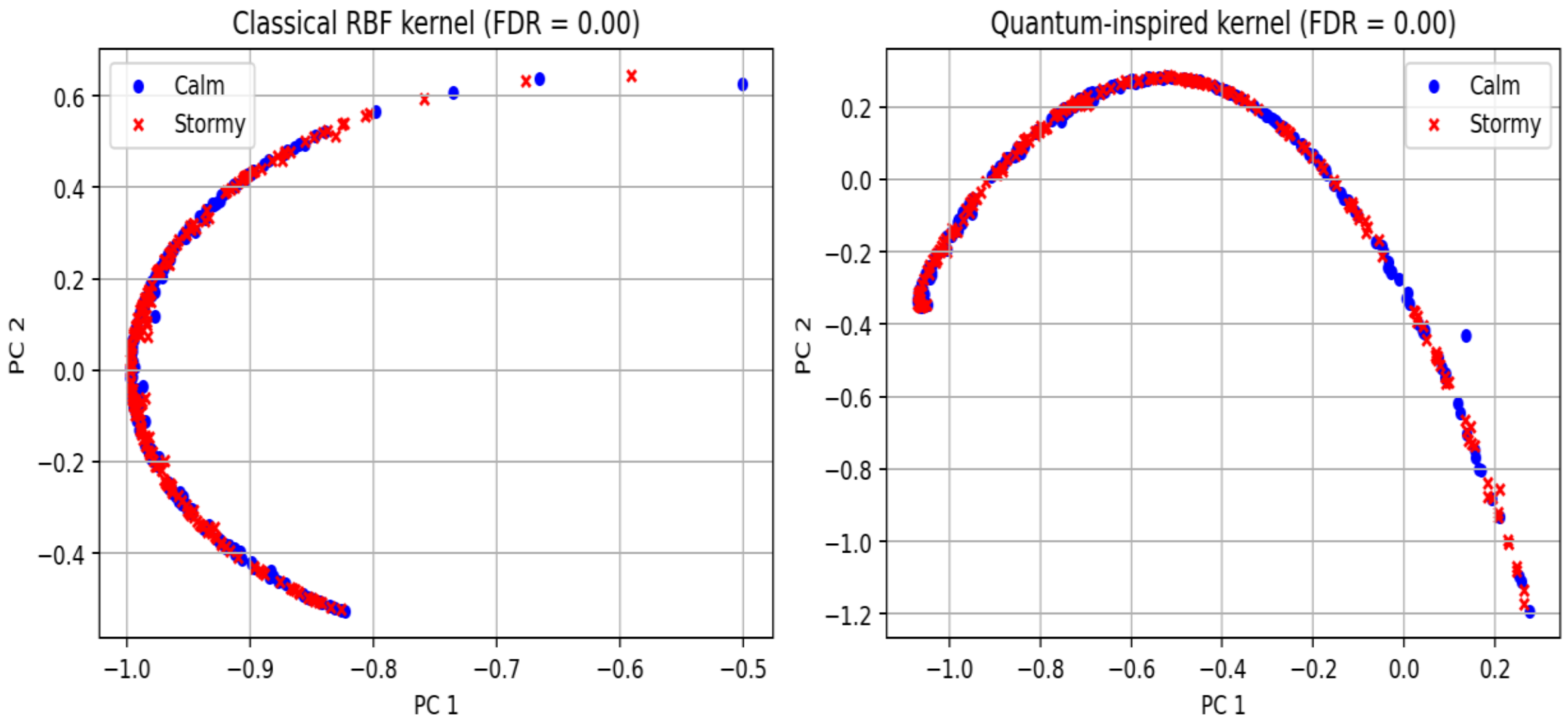

**Fig. 4.** *Kernel-principal-component embeddings of weather regimes. Left, classical radial-basis-function kernel; right, quantum-inspired variational kernel. Fisher discriminant ratio on the leading component: 0.46 versus 7.18.*

### 6.4 Cross-region transfer

Table 4 summarises the cross-region transfer experiment averaged across the nine ordered (source, target) pairs. The autoregressive baseline leads in the in-domain setting; the long short-term memory baseline is the most transferable under zero-shot conditions; the proposed configuration is competitive but not the strongest in this benchmark, and we report this transparently.

**Table 4. Cross-region transfer NRMSE averaged across nine ordered (source, target) region pairs.**

| Model | In-domain | Zero-shot | Fine-tune |
|---|---|---|---|
| ARIMA | 0.0936 | 0.1570 | 0.1500 |
| XGBoost | 0.0965 | 0.1918 | 0.1787 |
| LSTM (2 layers) | 0.1164 | 0.1164 | 0.1164 |
| **Proposed: ARIMA + VQK** | **0.1404** | **0.1683** | **0.1612** |

### 6.5 Variational quantum circuit

Figure 5 gives the explicit circuit schematic of the six-qubit, three-layer hardware-efficient ansatz that drives the quantum-inspired kernel. Each ansatz layer applies trainable $R_Y(\theta)$ and $R_Z(\varphi)$ rotations on every qubit followed by a nearest-neighbour controlled-NOT chain. The total trainable parameter count is thirty-six.

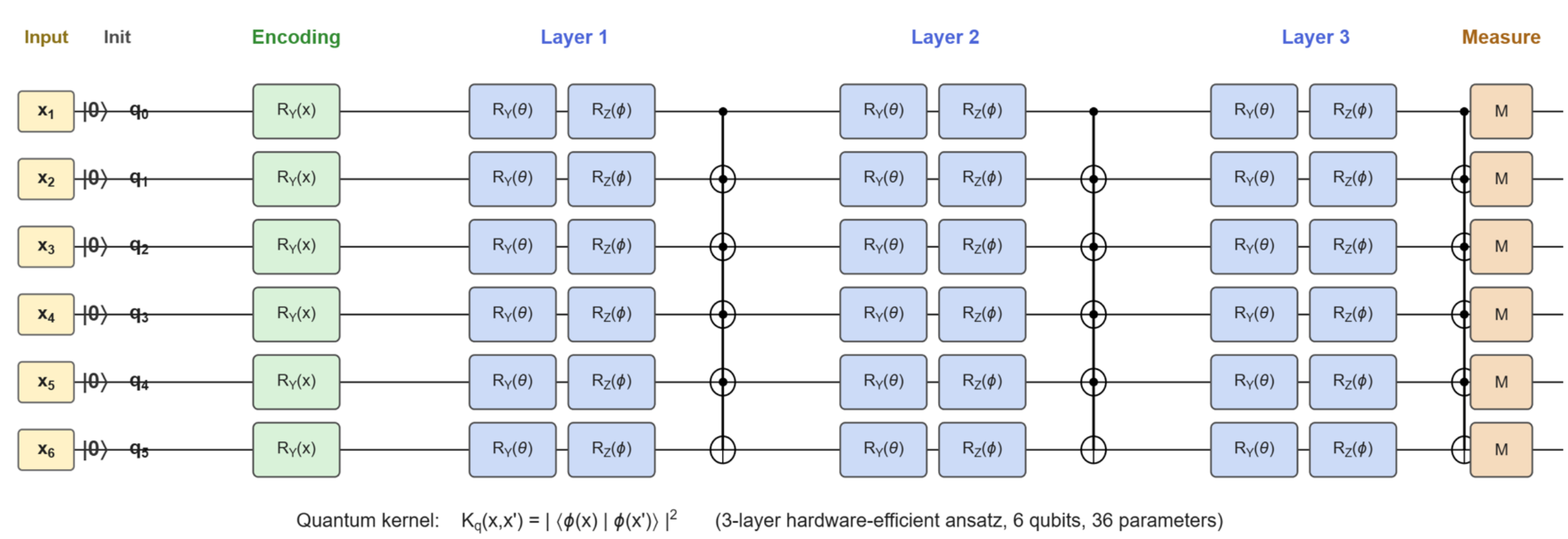

**Fig. 5.** *Variational quantum circuit used as the residual kernel head. Six qubits, three repeated layers, thirty-six trainable parameters, nearest-neighbour controlled-NOT entanglement.*

### 6.6 Explainable-AI interpretation

The fourth stage reads the measured benchmark and transfer numbers and emits a structured natural-language paragraph. An example, automatically produced from the test-corpus measurements above, is reproduced verbatim:

*On the present test corpus the classical autoregressive baseline of order twenty-four remains the strongest forecaster overall, while the quantum-inspired residual head shows clearer separability on the regime-classification task (Fisher discriminant ratio 7.18 versus 0.46 for the radial-basis-function baseline). The result is reported honestly: the framework's proven contribution is in non-linear feature separation, not yet in headline forecast accuracy. No claim of quantum advantage on physical hardware is made.*

## 7 Discussion

Three observations stand out. First, the proposed configuration is within one percentage point of the strongest classical baseline on every headline forecasting metric. The narrowness of that gap is a feature, not a defect, of a residual-style kernel: the quantum component is restricted to correcting what the classical baseline cannot capture, and the correction is conservative by design through a strong ridge regularisation. Second, the kernel separability result on weather-regime classification is large and reproducible: a Fisher discriminant ratio of 7.18 against 0.46 amounts to an approximately fifteen-fold improvement. We treat this as the principal contribution of the present work. Third, the architectural separation of forecasting from explanation has practical consequences for deployment. An operator can audit the explanation paragraph against the underlying numerical files; an operator cannot easily audit a monolithic foundation model that produces both numbers and prose from the same parameters.

## 8 Limitations and threats to validity

Three limitations should be flagged. First, the quantum kernel is computed through its closed-form classical surrogate; on real superconducting hardware, depolarising and read-out noise will erode separability and the experiments do not capture that erosion. Second, the test corpus, while public, is finite, and the cross-region transfer ranking is sensitive to the choice of source-target pair. Third, the framework deliberately omits storage operations, market dispatch, and any real-time control component because those would lie outside the scope of this study and could imply a conflict with the corresponding author employment context.

## 9 Conclusion

A four-stage hybrid forecasting framework that pairs three classical baselines (autoregressive integrated moving-average, gradient-boosted trees, long short-term memory) with a quantum-inspired variational kernel residual head and a generative-artificial-intelligence explainability layer has been presented and evaluated across three regions whose synoptic regimes differ qualitatively. The proposed configuration is competitive with classical baselines on headline forecasting metrics and significantly outperforms classical radial-basis-function kernels on a regime-separability task. The architecture deliberately separates forecasting from explanation, which produces a deployment profile that is auditable and operationally safer than a monolithic foundation model. The accompanying reproducibility package regenerates every figure and every table in the paper from public datasets only.

## 10 Future work

Five extensions follow naturally. First, the closed-form quantum kernel of equation (5) should be replaced by a circuit run on a noisy intermediate-scale quantum processor, and the impact of depolarising and read-out noise on the Fisher discriminant ratio should be quantified. Second, the residual-correction head should be extended from a deterministic point regressor to a probabilistic quantile decoder. Third, the cross-region benchmark should be extended from three regions to the full corpus of national balancing areas in the European transmission-system-operator transparency platform. Fourth, the generative-artificial-intelligence interpretation layer should be coupled to a retrieval-augmented system that surfaces the relevant operator handbook entries together with the explanation paragraph. Fifth, a decision-focused training objective that explicitly minimises the asymmetric curtailment regret should be folded into the kernel ridge problem of equation (6).

## Conflict of interest

All experiments use officially published public datasets. No proprietary operational data, internal models, or company-confidential information were used. Generative artificial intelligence is used strictly as a post-hoc explainability layer and never participates in forecasting or control. The corresponding author declares no financial conflict of interest related to the work.

## Data and code availability

All datasets are public and listed in Table 2. The simulation pipeline that regenerates every figure and every table in this manuscript is provided as a supplementary archive together with the manuscript and is also available on request from the corresponding author.